\documentclass[runningheads]{llncs}

\usepackage{eccv}
\usepackage{eccvabbrv}
\usepackage{graphicx}
\usepackage{amsmath,amssymb}
\usepackage{booktabs}
\usepackage{hyperref}

\title{Key-Frame Reasoning with SAM3: Third Place Solution for the MeViS-Text Track of the 8th LSVOS Challenge}
\author{Ce Bian\inst{1} \and Xusheng He\inst{1} \and Jinrong Zhang\inst{1} \and Canyang Wu\inst{1} \and Xianjing Han\inst{2} \and Jianlong Wu\inst{1,3}}
\titlerunning{Key-Frame Reasoning with SAM3}
\authorrunning{C. Bian \etal}
\institute{Harbin Institute of Technology, Shenzhen \and Nanyang Technological University \and Shenzhen Loop Area Institute \\
Team: HITsz-Dragon}

\begin{document}
\maketitle

\begin{abstract}
This report presents a two-stage, training-free solution for the MeViS-Text track of the 8th LSVOS Challenge. The task requires a model to localize and segment the object specified by a natural-language expression throughout a video. Such expressions often depend on temporal cues, including actions, interactions, directions, and relative positions. Our first stage uses Gemini-3.1 Pro via API to decompose a video-level event into instance-level targets, select a key frame for each target, and generate a discriminative description aligned with that frame. In the second stage, SAM3-agent produces a pixel-level seed mask on the selected frame, and the SAM3 video tracker propagates the mask bidirectionally through the video. Valid instances are grounded and propagated independently before their frame-wise masks are merged. All local SAM3 processing runs on a single NVIDIA GeForce RTX 4090 without task-specific training or model ensembling. Our method ranked third on the challenge test set, obtaining J\&F, J, F, N-acc., T-acc., and Final scores of 0.761, 0.7367, 0.7852, 0.8333, 0.9755, and 0.856593, respectively.
\end{abstract}

\section{Introduction}
Video object segmentation (VOS) aims to segment target regions throughout a video while maintaining temporal consistency under appearance changes, occlusions, and background distractors~\cite{davis,youtubevos}. Referring video object segmentation (RVOS) further requires the target to be determined from a natural-language expression~\cite{mevis_iccv,ding2025mevis}. RVOS is therefore a joint problem of visual grounding, language understanding, temporal reasoning, and pixel-level mask prediction.

The MeViS-Text track focuses on motion-centric referring expressions. A target can be specified through its motion direction, action, interaction with another object, relative position, state, or an exclusion relation. The model must compare observations across time to determine whether an instance truly satisfies the expression rather than relying only on its category or appearance in a single frame. Occlusions, viewpoint changes, cluttered backgrounds, and similar-looking instances further complicate both target identification and temporal segmentation.

A practical strategy is to interpret the expression and video content with a vision-language model before applying a segmentation or tracking model~\cite{refer_agent}. However, the temporal information in a complete motion expression may not align directly with a static frame. In addition, special tokens used as intermediate representations can weaken fine-grained instance attributes and spatial relations~\cite{one_token}. The central challenge is thus to turn a video-level event into a clear and executable image-grounding task. Prompt-conditioned video segmentation models provide a practical basis for the propagation stage: SAM2 established a general framework for propagating prompted masks across videos~\cite{ravi2024sam2}, while SAM3 further supports concept-conditioned segmentation and tracking~\cite{carion2026sam}.

We address this challenge with a two-stage training-free framework. Gemini-3.1 Pro identifies valid instances at the video level, selects the clearest key frame for each instance, and produces a discriminative description matched to that frame. SAM3-agent then generates a pixel-level seed mask, which is propagated through the video by the SAM3 tracker~\cite{carion2026sam}. Multiple valid instances are processed independently and merged only after propagation.

Our main contributions are as follows:
\begin{enumerate}
    \item We develop a two-stage, training-free RVOS framework that connects video-level event understanding with pixel-level video segmentation.
    \item We use Gemini-3.1 Pro to decompose complex events into instance-level targets and jointly determine key frames and discriminative descriptions, reducing ambiguity among similar instances.
    \item We generate pixel-level seed masks with SAM3-agent and propagate them bidirectionally with the SAM3 video tracker. Gemini-3.1 Pro is accessed via API, and all local SAM3 processing runs on a single RTX 4090. The method ranks third in the challenge.
\end{enumerate}

\section{Solution}
\label{sec:solution}

\subsection{Overall Framework}
\label{sec:overall}

We adopt a two-stage framework. The first stage performs event understanding, while the second stage comprises single-frame localization and video propagation. First, we use a multimodal large language model to analyze the video and its motion event expression, decomposing the original query into one or more instance-level targets. For each target, the model selects the key frame most suitable for localization and generates a discriminative description conditioned on the visual content of that frame. The key frame and its description are then fed into a SAM3-agent, which directly produces a pixel-level seed mask, and the SAM3 video tracker propagates this mask across the entire video. For expressions that contain multiple valid instances, each instance is localized and tracked separately, and the resulting predictions are merged into an event-level result.

This design decouples cross-frame event understanding from pixel-level segmentation. Motion expressions typically encode information about actions, directions, interactions, and relative positions, which is difficult to use directly as a single-frame segmentation prompt. After the first stage, each target has a well-defined instance identity, an unambiguous reference frame, and a description that matches the corresponding frame, providing more stable inputs for the subsequent segmentation stage. Meanwhile, the second stage directly produces a seed mask instead of relying on bounding boxes as the primary intermediate representation, so that object contours and spatial details can be preserved more completely for video propagation.

\subsection{Two-Stage Processing Pipeline}
\label{sec:pipeline}

\subsubsection{Stage 1: Event Decomposition via Key-Frame Reasoning}
\label{sec:stage1}

For each video and its target event, we use Gemini-3.1 Pro to analyze the video content holistically. This stage does not simply restate the event; instead, it identifies the target instances referred to by the expression and organizes them into independently processable records. Other objects that merely help describe actions or relations in the expression are treated only as localization cues and are not included in the final segmentation targets. If multiple distinct physical instances satisfy the event, a separate record is generated for each; if the same instance appears at multiple moments, it is processed only once.

After instance identification, the model selects, for each target, the key frame most suitable for single-frame localization. The selection primarily considers whether the target is clearly visible, whether its contour is complete, whether occlusion is low, and whether it can be distinguished from similar instances in the frame. For longer videos, we uniformly sample the input frames and record the correspondence between the sampling order and the original frame indices, so that the positions returned by the model can be mapped back accurately to the real video frames.

Then, based on the selected key frame, Gemini-3.1 Pro generates a discriminative description for each instance. The description includes not only the target category but also the visible attributes of the target in the current frame, such as color, texture, shape, and clothing, as well as its spatial relations to surrounding objects or image regions. For multiple instances of the same category, the model describes them separately using different appearance and location cues, ensuring, as much as possible, that each description points to exactly one target within the corresponding key frame. Through this process, motion expressions that originally rely on cross-frame behavioral understanding are converted into a set of image-localization tasks with clearly defined targets, key frames, and dedicated descriptions.

The first stage stores the name, key frame, and discriminative description of each instance in a structured form. If the model determines that no target satisfying the event exists in the video, it returns an empty result. This output is passed directly to the second stage, so that the SAM3-agent does not need to re-understand the full motion event and can focus on instance-level localization within the key frame.

\subsubsection{Stage 2: SAM3-agent Localization and Temporal Propagation}
\label{sec:stage2}

Given the key frames and discriminative descriptions, we run a SAM3-agent on each target instance separately. This process accomplishes single-frame target localization through multi-round interaction between a multimodal large language model and the SAM3 tool. In our implementation, Gemini-3.1 Pro is responsible for analyzing the key frame and the instance description, planning tool calls, and judging the segmentation results returned by SAM3. Specifically, Gemini-3.1 Pro first generates a text phrase suited to SAM3 based on the target's characteristics, and SAM3 produces candidate masks accordingly. Gemini-3.1 Pro then checks the candidate's identity, spatial location, and mask completeness against the original description, and decides the next action.

If the current result fails to localize the target accurately, Gemini-3.1 Pro further inspects the candidate mask, or adjusts the text phrase and invokes SAM3 again, until a satisfactory result is obtained. The phrase used to call SAM3 may be more concise than the initial discriminative description, but the judgment of the candidate mask always relies on the initial description, thereby avoiding a shift in the referred object caused by prompt simplification. When a candidate mask covers the target completely and contains no clearly irrelevant regions, the SAM3-agent confirms it as the seed mask; if no matching target exists in the key frame, it returns an empty mask. Through this multi-round interaction, the SAM3-agent progressively eliminates candidates that do not match the description and mitigates the localization ambiguity caused by similar instances.

Because a bounding box can only roughly enclose the target and tends to include irrelevant regions when targets are in contact, the background is complex, or similar instances are present, we directly use the pixel-level mask output by the SAM3-agent to initialize the video tracker. A pixel-level seed mask provides more accurate object contours and spatial positions, offering a more reliable initial prompt for subsequent temporal propagation.

\begin{figure*}[t]
    \centering
    \includegraphics[width=\textwidth]{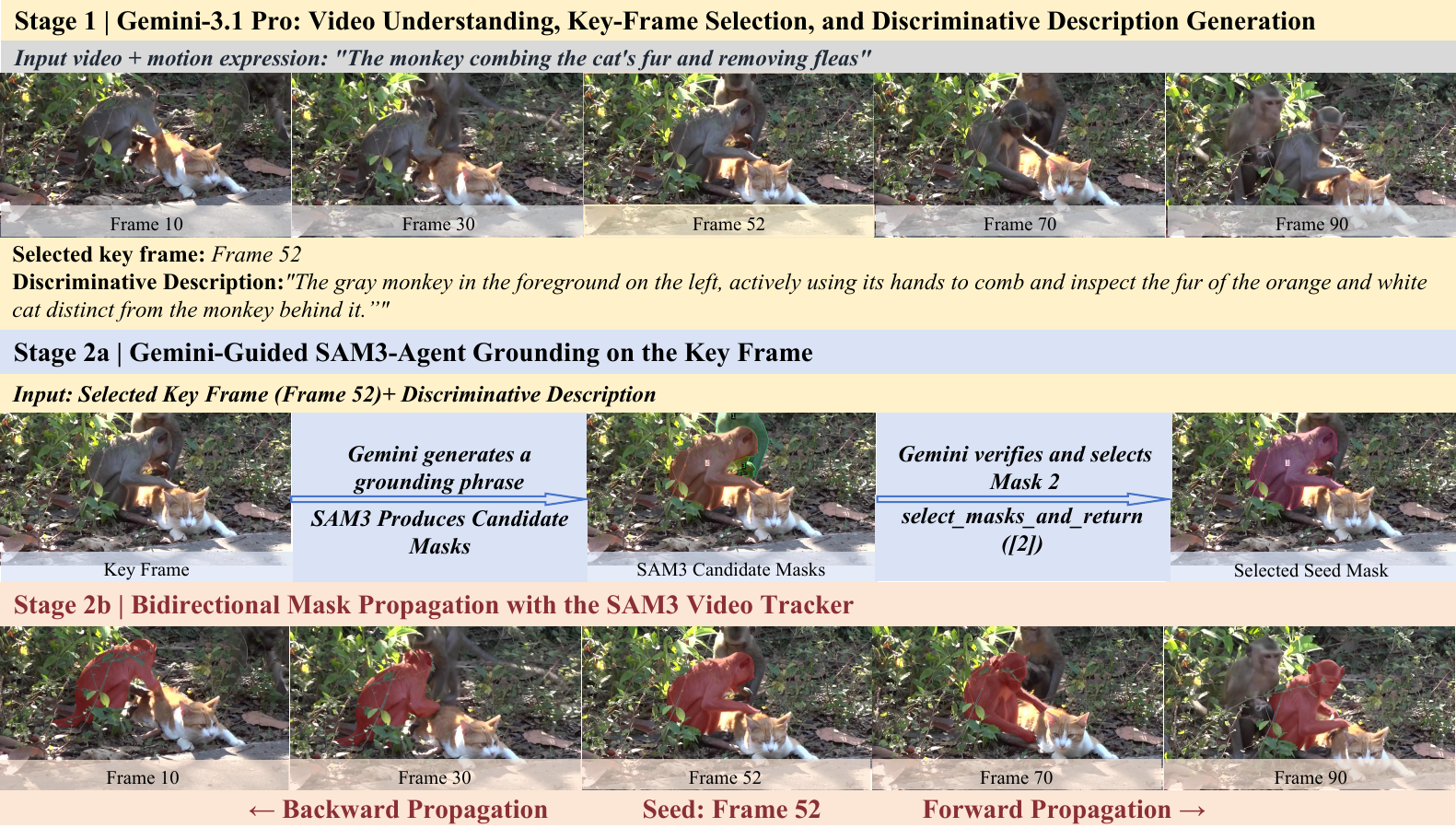}
    \caption{Overview of our complete two-stage pipeline on a representative MeViS-Text example, from key-frame reasoning and discriminative-description generation to SAM3-agent grounding and bidirectional mask propagation.}
    \label{fig:qualitative}
\end{figure*}

Once the seed mask is obtained, we initialize the SAM3 video tracker on the corresponding key frame and propagate it toward both the beginning and the end of the video, thereby covering the complete sequence before and after the key frame. The tracking results are binarized and saved as per-frame masks that correspond one-to-one to the original video frames. Figure~\ref{fig:qualitative} shows a representative example of the whole process from key-frame reasoning to full-video segmentation.

\newpage
For expressions containing multiple valid instances, we propagate each instance independently according to its key frame, discriminative description, and seed mask, and then merge the instance masks within the same frame to form the final event-level prediction. This reduces mutual interference among different targets during initialization and allows each instance to be tracked from its clearest moment.

\section{Experiments}
\subsection{Challenge Description}
We participated in the MeViS-Text track of the 8th Large-scale Video Object Segmentation (LSVOS) Challenge~\cite{lsvos2026}. Given a video and a natural-language expression, the system must output a binary PNG mask for every original video frame. Expressions may specify a simple object or impose constraints involving motion, interaction, direction, relative position, or state. The task consequently requires both referent identification and temporally consistent segmentation.

\subsection{Implementation Details}
The method is used only for inference: no task-specific training or model ensemble is employed. Gemini-3.1 Pro is accessed via API for event analysis, instance decomposition, key-frame selection, discriminative-description generation, tool planning, and candidate-mask evaluation. SAM3 generates candidate masks on key frames, and its video tracker performs temporal propagation.

Local SAM3 inference and video tracking use one NVIDIA GeForce RTX 4090. We store submission masks as \path{<video_id>/<expression_id>/<frame>.png}. Missing predictions are replaced with all-zero masks so that every expression has an output mask for every input frame.

\subsection{Evaluation Metrics}
Jaccard index (J) measures region overlap between a prediction and its ground truth, while boundary F-score (F) evaluates boundary accuracy. J\&F is their mean. N-acc. evaluates no-target expressions, and T-acc. evaluates target expressions. The Final score combines J\&F, N-acc., and T-acc.

\subsection{Quantitative Results}
Our method ranks third on the official MeViS-Text test set with a Final score of 0.856593. It obtains a T-acc. of 0.9755, J\&F of 0.761, J of 0.7367, F of 0.7852, and N-acc. of 0.8333. Table~\ref{tab:results} reports the top three entries. With Gemini-3.1 Pro accessed via API, the local SAM3 components run on a single RTX 4090 without task-specific training or ensembling. These results demonstrate the feasibility of the two-stage pipeline.

\begin{table}[t]
\centering
\caption{Top-three official MeViS-Text test-set results. The proposed method is shown in bold.}
\label{tab:results}
\scriptsize
\setlength{\tabcolsep}{2.5pt}
\begin{tabular}{c c c c c c c}
\toprule
 Rank & J\&F & J & F & N-acc. & T-acc. & Final \\
 \midrule
1 & 0.7922 & 0.7657 & 0.8187 & 0.9444 & 0.9877 & 0.908134 \\
2 & 0.7163 & 0.6894 & 0.7432 & 0.9722 & 0.9755 & 0.888004 \\
3 & \textbf{0.761} & \textbf{0.7367} & \textbf{0.7852} & \textbf{0.8333} & \textbf{0.9755} & \textbf{0.856593} \\
\bottomrule
\end{tabular}
\end{table}

\subsection{Qualitative Results and Discussion}
Figure~\ref{fig:qualitative} shows an example for the expression ``The monkey combing the cat's fur and removing fleas.'' The video contains a foreground monkey, another monkey behind it, and a cat. Stage 1 selects Frame 52 and generates a description that distinguishes the foreground target from the other monkey. SAM3-agent produces candidate masks using the phrase ``monkey'' and selects the target seed mask according to the full description. The SAM3 video tracker then propagates the mask throughout the video.

The example illustrates how a key frame and a discriminative description can separate similar instances and provide a clear spatial initialization for propagation. The method nevertheless depends on the first-stage decision: a missed instance or unsuitable key frame affects later grounding and tracking. A single key frame can also be insufficient for expressions dominated by complex directional, negative, or behavioral constraints. No-target expressions remain a relative weakness.

\section{Conclusion}
We presented a two-stage, training-free approach for motion-centric RVOS in the MeViS-Text track. Gemini-3.1 Pro performs instance-level event decomposition, key-frame selection, and discriminative-description generation. SAM3-agent then produces pixel-level seed masks, which the SAM3 video tracker propagates bidirectionally through the video. Independently tracked instances are merged into the final prediction.

Gemini-3.1 Pro is accessed via API, while the local SAM3 components run on a single NVIDIA GeForce RTX 4090 without task-specific training or model ensembling. The method ranks third in the 8th LSVOS Challenge MeViS-Text track, achieving J\&F of 0.761 and a Final score of 0.856593. The results support the use of strong multimodal event understanding together with pixel-level grounding and temporal propagation for motion-centric RVOS.

\bibliographystyle{splncs04}
\bibliography{main}

\end{document}